# Predicting challenge moments from students' discourse: A comparison of GPT-4 to two traditional natural language processing approaches

Wannapon Suraworachet, Jennifer Seon, Mutlu Cukurova
University College London, United Kingdom
wannapon.suraworachet.20@ucl.ac.uk

## ABSTRACT

Effective collaboration requires groups to strategically regulate themselves to overcome challenges. Research has shown that groups may fail to regulate due to differences in members' perceptions of challenges which may benefit from external support. In this study, we investigated the potential of leveraging three distinct natural language processing models: an expert knowledge rule-based model, a supervised machine learning (ML) model and a Large Language model (LLM), in challenge detection and challenge dimension identification (cognitive, metacognitive, emotional and technical/other challenges) from student discourse, was investigated. The results show that the supervised ML and the LLM approaches performed considerably well in both tasks, in contrast to the rule-based approach, whose efficacy heavily relies on the engineered features by experts. The paper provides an extensive discussion of the three approaches' performance for automated detection and support of students' challenge moments in collaborative learning activities. It argues that, although LLMs provide many advantages, they are unlikely to be the panacea to issues of the detection and feedback provision of socially shared regulation of learning due to their lack of reliability, as well as issues of validity evaluation, privacy and confabulation. We conclude the paper with a discussion on additional considerations, including model transparency to explore feasible and meaningful analytical feedback for students and educators using LLMs.

**ACM Reference Format:**

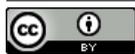

## 1 INTRODUCTION

Collaboration has been highlighted as a key twenty-first-century skill, and the increased need to think and work together across disciplines and sectors has only accelerated interdependence [2, 37, 54]. Broadly speaking, collaborative learning (CL) involves learners interacting in smaller groups to construct a shared understanding in an attempt to solve problems or achieve shared goals that would otherwise be difficult to accomplish alone [10, 28, 44, 47]. It requires mutual effort by each group member to negotiate a shared understanding in order to achieve a common goal [13, 28]. CL provides a community of status equals – peers – in which students learn the "skill and partnership" [5, p.642] to co-construct knowledge in a state of continual negotiation, not only for academic purposes but also for their professional and daily lives.

Successful collaboration requires more than collegiality or collective sameness; it requires group members to "approach one another as equals and work out concepts through the cogeneration and consensual validation of intellectual strategies. They learn from one another not by copying or adopting the other's competence ... but by mutually devising plans together in a collaborative effort" [1, p.334]. In essence, collaboration requires learners to engage in regulation of learning (ROL) processes, which involves the collective planning, monitoring, evaluating, and taking control of their own learning to achieve set goals [21]. As the group navigates through the collaborative task, members can individually regulate themselves (self-regulated learning, SRL), co-regulate others (co-regulation, Co-RL), and/or collectively regulate as a group (socially shared regulation, SSRL) to overcome the wide range of challenges that may arise [23, 31].

For regulation, group members must first acknowledge the challenge moments that might hinder effective collaboration and develop suitable strategies to overcome these challenges together [29, 35]. The use of ROL strategies has been linked to higher academic outcomes [6, 45], with research showing groups highly engaged in SSRL processes negotiated shared task perceptions, goals, plans, and strategies [23, 35] while maintaining positive socio-emotional interactions, or establishing "mutual trust" to overcome challenges collectively [18]. Without awareness of one's own SRL and where the group stands as a whole, various perceptions of the challenges can lead to inappropriate or misaligned regulatory processes by group members [20, 28]. However, due to differences in personal socio-historical experiences situated in various contexts [23], recognising challenge moments and aligning task perceptions and goals to coordinate SSRL strategies may require external regulatory support [28]. Research has shown groups that fail to accurately perceive challenge moments tend to activate procedural and behavioural (i.e., "routine-level") strategies, while highly regulated learners focus more on deep-level processes such as cognitive and metacognitive strategies to collectively overcome challenges [27, 35].

Although researchers have begun to develop computer-based pedagogical tools or pedagogical agents to support SRL [3, 41], gathering SRL data through macro-level self-reports may not accurately reflect SRL strategies [49]. Such global measures also limit the study and support of regulation processes "on the fly" [55], hindering real time and ongoing support for regulation in CL. This study aims to address the lack of research that detects opportunities for regulation to take place by automatically identifying the type of challenges faced by group members through trace data (audio transcriptions) to overcome the concerns of validity and a lack of fine-grained information in self-reports. An accurate identification of challenge moments is a crucial prerequisite to the deployment of appropriate ROL strategies which would support the alignment of strategies amongst group members for successful collaboration [28].

## 2 RELATED WORKS

### 2.1 Challenge moments in collaboration and SSRL

Previous studies have explored SSRL strategies for motivation and emotion in which students were found to adapt their strategies to fit the specific situated challenges [26, 30]. Both studies have highlighted the dynamic nature of SSRL processes to respond to the situated circumstances as they continually evolve and change in response to the interactions between group members. Järvelä et al. [27] focused on the types and patterns of regulation emerging over time in CL and found three kinds of regulation profiles (strong, progressive, weak) which used differing SSRL strategies (deep-level vs. routine-level). Malmberg et al. [35] found supporting evidence with high-performing groups progressing their SSRL over time, as their challenges and corresponding regulation strategies varied temporally, while the low-performing groups remained focused on activating routine-level strategies, if any at all. Although these studies have made important contributions to the field, manual codings were used to analyse data which are not conducive to real-time feedback. This study aims to explore the potential of automating the coding process to address the need for "on the fly" regulation.

Despite a plethora of SSRL research focusing on analysing regulatory strategies, only a minority of them have been contextualised by students' encountered challenges. Hadwin, Järvelä and Miller [23] emphasised the need to identify challenging episodes, defined as a situation when learners individually or collectively encounter difficulties when investigating regulatory processes in collaboration [23]. They reasoned that challenges help frame goals and plan for regulation which, in turn, helps understand why such regulatory strategies were used in the contexts. This also provides opportunities to detect whether any regulation of challenge is followed and if so, whether the deployed regulatory strategies were successful or correspond to a problem at hand. Although previous studies have made important contributions to the field of collaborative learning, manual coding was used to analyse data in all, which is not conducive to real-time feedback. This study aims to explore the potential of automating the coding process to address the need for "on the fly" regulation.

### 2.2 Learning analytics and natural language processing for the detection and support of challenge moments in collaboration

Learning analytics (LA) is an emerging field that leverages available educational data about learners, interactions, and contexts for enhancing learning processes [32]. LA undergoes a cycle of collecting, distilling, and representing extracted insights to inform decisions of educators and learners [46]. The education field benefits from LA through automatic data processing with machine learning (ML) techniques and data-driven feedback to close the feedback loop. Typically, educational qualitative data is manually processed which is labour-intensive and difficult to scale for timely feedback. Advance in natural language processing (NLP) helps computationally recognised patterns in texts [14], for instance, group discourse which is pervasively available in collaborative settings.

Multiple NLP techniques have been used in extracting features from discourse. The conventional approach involves the use of bags-of-words [1] in analysing texts by considering frequencies of words or consecutive n-words (n-grams) presented in texts. These word-based representations can then be used as key features to further analyse a downstream task such as identification of coherence discussion topics based on n-grams [50]. In addition to local word representation, there is also a more universal word representation through a dictionary that comprises a predefined list of words representing targeted constructs. To illustrate, LIWC (Linguistic Inquiry and Word Count), a commercial dictionary contains representative words of psychological processes and has been widely used in educational research [40]. Apart from word-based representations, linguistic structures of sentences such as part-of-speech (POS) categories of words have also been considered in the NLP approach. For instance, Sullivan and Keith [52] employed parts of speech in combination with trigrams to analyse spoken dialogues of middle school students engaged in a robotics challenge. These POS trigrams were then mapped with actions and objects in computational environments to infer problem-solving processes. Emara et al. [15] similarly leveraged POS tri-grams obtained from group discourse to identify group patterns of regulation in collaborative open-ended problem-solving contexts. Other approaches such as sentiment analysis help classify text into various sentiment categories such as positive, neutral or negative, or certain emotional labels such as happy or sad. A notable aspect of deploying sentiment analysis in educational contexts is to study how emotions or affective stages play roles in learning [19]. However, the extent to which these features will be useful in challenge identification has yet to be studied.

These NLP-based features can be effectively employed in conjunction with a range of methodologies for data analysis. This included a conventional rule-based approach in heuristically formulating decision rules by domain experts to more sophisticated ML techniques such as generating predictive models from training datasets (supervised ML). To illustrate, Pugh et al. [43] combined n-grams and LIWC features extracted from group transcription to build a random forest classifier to determine collaborative problem-solving facets. Zheng and colleagues [58] created a semi-auto ML pipeline combining human coding and NLP features to identify SRL vs SSRL constructs from group chats. While current research favours ML techniques, others argue that well-established expert-engineered features could outperform ML approaches in some tasks and conditions such as identifying student affect and disengagement [5].

In addition to existing techniques, there has been a growing recent interest in the development and utilisation of Large Language Models (LLMs), which are versatile artificial intelligence models trained on extensive datasets. Due to their pre-trained nature on large datasets, LLMs acquire a deep understanding of patterns in human language, enabling them to receive natural language prompts or instructions and respond accordingly [33]. This presents educators, even those lacking a programming background, with opportunities to harness these models for their specific purposes. Notably, LLMs have gathered recognition for their effectiveness and capacity to generalise across tasks, even when confronted with previously unseen data (zero-shot learning) or when provided with minimal training (one-shot or few-shot learning). Prominent LLMs include GPT (Generative Pre-trained Transformer) developed by OpenAI [39] and BERT (Bidirectional Encoder Representations from Transformers) developed by Google [12]. There has been growing interest in applying LLMs in educational contexts. For example, Ma, Celepkolu, and Boyer [34] utilized LLMs in detecting impasse or group conflicts due to differences in opinions or insufficient ideas which could hinder collaboration progress. They proposed multimodal modelling techniques including analysis of group dialogue using a BERT-based model to detect three categories of impasse (impasse disagreement, impasse insufficient ideas and non-impasse). Their results show that speaker-embedded linguistic features could potentially indicate an impasse. Even though there are conceptual overlaps between impasse and challenge that both impede success in collaboration, impasse only covered socio-cognitive aspects of challenges, identified in this study. Additionally, their study concerns the dyads dialogue in which their patterns may differ for a bigger group of students. Moreover, they only identified three impasse categories which might not be meaningful to provide support.

Despite the potential of NLP and recent developments in analytic techniques, there are not sufficient studies and pragmatic implementations of NLP in SSRL research. While some studies applied NLP to unravel the regulatory process [15, 58], none of them target the emergence of the regulatory process, i.e., predicting challenges and identifying challenge dimensions. We propose a first step in providing analytics regulatory support through the detection of challenges and their dimension in collaborative discourse using NLP approaches. Three different NLP modelling approaches, namely a rule-based, supervised ML, and LLM approach were compared in terms of their performance. The following are two main research questions that will be covered in this study:

● RQ1: How do performances of three different NLP modelling techniques (rule-based approach, supervised machine learning, and GPT-4) differ in predicting students' challenge moments in collaborative discourses?

● RQ2: What are the relative advantages and disadvantages of models for improving teachers' practice?

Reflecting upon the models' performance, we further discuss other advantages and disadvantages of the three models targeting specifically how they can be used in enhancing teaching and learning practices. To the best of our knowledge, this work is the first that focused on identifying challenges and challenge dimensions from collaborative discourse through an investigation of various NLP approaches including LLMs. This will contribute to a boarder field of regulation of learning research by illuminating the potential for automating challenge detection of students during collaboration.

## 3 METHODOLOGY

### 3.1 Contexts

This study involved 44 students in a semester-long postgraduate educational technology programme. Institutional ethical approval and participant consent were obtained before data collection. Students were grouped heterogeneously according to their backgrounds, gender, first language and years of working experience, with 4-5 members per group, and tasked with designing an educational technology solution over nine weeks. The group work was not part of the summative assessment but aimed to enhance content understanding through collaboration. The primary assessments were a critical essay and a weekly reflection on module engagement. Each week, students engaged in 60-minute collaborative tasks on the Miro[1] platform, integrating weekly module content into their final design. The students collaborated in-person, seated around a T-shaped table, with teachers present for assistance. As the study focused on verbal communication, group discussions were recorded using a conference microphone (Boya BY-MC2). Students also completed short pre/post-surveys weekly, assessing their motivation, preparation, goals, and collaborative experiences.

### 3.2 Data processing

This study incorporated qualified audio from 28 sessions. Transcriptions were generated from audio recordings using Whisper.ai[2], an open-source automatic speech recognition (ASR) system, providing start and stop speaking times and content. Speaker diarization was performed using the open-source Python library, Pyannote[3], which identified the start and stop time of speech with detected speakers. The maximum number of speakers was set to six to accommodate the group members and the teacher. Speakers and content were then aligned and merged based on speaking time. Due to inaccuracies in speaker detection and transcription, manual corrections were made by the first author. Given the large group size (4-5 students), multiple discussion threads may have been ongoing simultaneously. The manual correction focused on the most comprehensive thread, disregarding others. Video recordings were utilized during manual correction process to better understand the discussion context. Information about students' gestures and contexts was also noted to aid interpretation. Anonymisation was conducted using the spaCy[4] open-sourced library's named entity recognition (NER) model to filter identifiable information. The pre-trained NER model 'en_core_web_sm' was used to mask entities like names with their categories, e.g., 'Henry' was masked as <person>. However, names of education technologies and 'English' as a language were not masked for context understanding. Human screening and pseudonyms replacements were performed to ensure anonymity. Finally, speaking time and content were computed into utterances, defined as a single person's turn-talking.

---
[1] https://miro.com
[2] https://openai.com/research/whisper
[3] https://github.com/pyannote/pyannote-audio

Table 1: Final coding schemes of dimensions of challenges and their distribution

| Dimension (% in utterances, and episodes) | Sub-dimension | Example |
| --- | --- | --- |
| Cognitive challenges (C): struggling to comprehend (7.5%, 52%) | Expressing confusion either about the task, contents, or task expectation (C1) | "What's the meaning of...?", "I'm confused about ..." |
| | Expressing concerns over the brainstorming ideas (C2) | "Maybe the solution already existed.", "It may not work in other contexts." |
| | Questioning one's ability to communicate clearly (C3) | "Does it make sense?", "I'm not sure if this makes sense", "I am not sure that I have explained myself clearly." |
| | Struggling to understand peers in terms of the terminology used or the proposed ideas (C4) | "What do you mean by...?", "In terms of what?", "You mean to... or...?" |
| Metacognitive challenges (M): struggling to monitor, execute or control the task (3%, 36%) | Raising concerns over time/progress (M1) | "We have to finish it in.. minutes.", "we have 2 tasks left.", "Where are we?", "Shall we move on?" |
| | Expressing confusion when executing tasks in the brainstorming platform (M2) | "Who's blue? (blue refers to a blue sticky note in this context)", "What should we do?", "So we focus on one, two, or we do everything, right?" |
| | Expressing concerns over the strategies/approaches that the group used (M3) | "I don't think that is the right approach." |
| Emotional challenges (E): expressing negative feelings. (1%, 10%) | Expressing emotion or frustration. (E1) | "I'm lost.", "I'm confused.", "I'm not really happy.", "I'm not good at..." |
| | Expressing non-interests. (E2) | "I don't want to...", "I'm not motivated in this." |
| | Experiencing difficulty. (E3) | "That's hard.", "It's complex." |
| Technical/other challenges (T): experiencing external/ environmental/ technical challenges (1.8%, 18%) | The faced challenges were related to environmental situations, technical issues or personal circumstances which haven't been identified above. | "It's not working on iPad.", "I have a problem with Miro.", "I can't make it.", "How to add stickers?" |

### 3.3 Human coding of dimension(s) of challenges

Two educational researchers who are familiar with the module reviewed the literature on dimensions of challenges in collaboration, forming initial coding schemes compiled from multiple studies targeting challenges or the emergence of regulation [4, 22, 26, 35, 38, 53]. These schemes were refined according to the contexts through pilot coding, clarification discussions and conflicting resolving, resulting in final coding schemes represented in Table 1. The researchers agreed to code at the utterance level and code each challenge dimension separately due to their non-exclusivity. They re-coded a previous session and two randomly selected sessions, representing 10% overlapped coding which achieved moderate inter-rater reliability across the coding dimensions (Cohen's Kappa = 0.74, 0.71, 0.74 and 0.82 for cognitive, metacognitive, emotional and technical/other challenges, respectively). They also coded episodes of discussion which refer to multiple utterances of a single topic. Audio recordings were randomly assigned for independent coding, with off-topic discussions excluded. The final data consisted of 10799 clean utterances (average of 386 utterances per session) and 882 discussion episodes (average of 31.5 episodes per session). Table 1 also shows the prevalence of challenge dimensions.

## 4 NATURAL LANGUAGE PROCESSING TO MODEL STUDENTS' CHALLENGE MOMENTS

Predicting whether there is a challenge and determining its dimensions (cognitive, metacognitive, emotional or technical) is a multi-label classification problem since the dimensions of challenges are not mutually exclusive. We constructed a two-step approach to tackle this problem. First, a model predicting whether there is a challenge was built. Then, to further determine the challenge dimensions of the predicted data, the dimension-specific classifiers were trained. We hypothesised that this 2-step approach could increase the performance due to the dependency created to initially filter out 'no challenge' data which otherwise would lead to very unbalanced datasets. Three different approaches to predict whether there is a challenge and identify its dimensions of challenges were experimented with in this study: 1) a rule-based approach, 2) supervised ML and 3) a generative LLM (GPT-4). Figure 1 represents the overall workflow.

---
[4] https://spacy.io/models/en

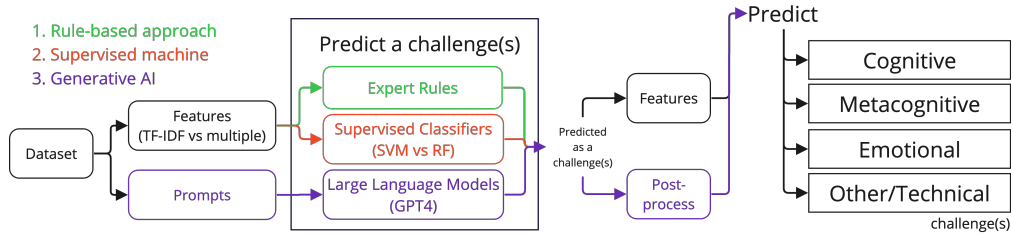

Figure 1: Pipelines of the three approaches

## 4.1 Feature extraction

In this section, the features and their description are discussed and summarised in Supplementary material S1[5].

*4.1.1 N-grams with TF-IDF.* The conventional Natural Language Processing (NLP) approach uses n-grams as features for classification, which are frequency counts of consecutive words in data. This can be combined with a term frequency-inverse document frequency (TF-IDF) weighting function to penalise common words and reward distinctive ones [36]. While unigrams produce a less sparse matrix and trigrams capture more relevant contexts for interpretation [15]. For example, when considering words in a trigram 'what pain point', it could be interpreted differently than the combined meaning of the three words. In this study, the generation of n-gram features involves tokenization, text preprocessing, stemming (transformation of words into their based form [48]), lemmatization, and removal of special characters and English stop words. The preprocessed texts are then forwarded to a TF-IDF vectorizer[6] to calculate and weigh the relative frequency of 1 to 3-grams.

*4.1.2 POS features.* Apart from n-gram features, we engineered part-of-speech (POS) of words as features. POS tags such as pronouns verbs and wh-adverbs have been promisingly used to identify regulation types from students' discourse [15]. We then explore the potentials of POS tags in identifying challenges. In the discussion, we observed that students could refer to their own situation (e.g., "I am confused.") or discuss in the contexts i.e., what is the challenge(s) that the group wants to solve. (e.g., "Students may be confused."). So, we hypothesised that the presence/absence of pronouns could help differentiate the two cases. POS tags could be identified from lexical resources e.g., a corpus containing labelled POS tags, or a statistical model predicting POS. In this study, we utilised NLTK pos_tag[7], an open-source Python library to perform POS tagging.

*4.1.3 Dialogue Acts.* In addition to typical NLP features, we also explored public datasets that have potential labels to be experimented with. Dialogue acts are other appealing features for our study. The identification of dialogue acts has been recognised as an important step in discourse analysis to understand the roles of utterances and speakers' intentions in communication [51]. To meet this end, we used the NPS Chat corpus[8] which is a part of the open-sourced NLTK package that contains approximately 10k online chat posts with labelled dialogue-act tags. In this NPS Chat corpus, the posts were categorised into one of fifteen dialogue-act tags such as Emotion, Emphasis, Greet, Reject, Statement, etc. [17]. According to our observation, Rejection, Wh-question and Yes/no questions could be related to an expression of challenges. For example, "What should we do?" which refers to a challenge of students in managing their task, should be classified as 'Wh-question'. Therefore, the three dialogue-act tags were included as potential features. We built a Naive Bayes classifier from the NPS Chat corpus to determine whether there is a question (Wh or yes/no questions) or rejection presented in utterances/episodes. In addition to features extracted from word occurrences, grammatical roles of words could also be useful in text analysis.

*4.1.4 Sentiment analysis.* Sentiment analysis is another longstanding and prominent area of NLP research in determining positive, neutral or negative sentiment from a given text. This is particularly related to emotions expressed during collaboration [29]. VADER (Valence Aware Dictionary and sEntiment Reasoner), an open-sourced Python package for sentiment analysis[9] was explored in this study. In general, it is a dictionary and rule-based package targeting sentiments in social media content, considering both word presence and order to determine degrees of intensifiers. It produces a normalised compound score for a sentence, with -1 indicating extreme negative sentiment and +1 indicating extreme positive sentiment. The developers suggest a threshold of -0.05 for detecting negative sentiment [25], which we adopted in our study.

*4.1.5 GoEmotions.* GoEmotions, a large-scale annotated dataset for emotional classification tasks, comprises 58k Reddit comments manually classified into 27 non-mutually exclusive emotional categories [11]. These categories, derived from psychology literature, were revised to fit the dataset and hierarchically clustered into positive, ambiguous, and negative sentiments. A logistic regression classifier was built from TF-IDF features for each emotion category using an undersampling technique. Resonating with cognitive challenges concepts, we chose 'curiosity' and 'confusion' as additional features. As recognised in SSRL research that emotion challenges are tightly coupled with negative feelings or frustration expressed during collaboration [38], we hypothesised that emotional challenges might be related to negative sentiment categories. All negative sentiment groups such as 'anger', 'annoyance', and 'disappointment' were considered as features.

---
[5]https://osf.io/ta87g/?view_only$=$74d86897cf1a4b02808c5821221a3e42
[6]https://scikit-learn.org/stable/modules/generated/sklearn.feature_extraction.text.TfidfVectorizer.html
[7]https://www.nltk.org/api/nltk.tag.pos_tag.html
[8]http://faculty.nps.edu/cmartell/npschat.htm
[9]https://github.com/cjhutto/vaderSentiment

*4.1.6 Post-survey responses.* Apart from engineering features from public datasets, weekly post-surveys were explored. Adapted from [4], one of the post-survey questions ("Throughout the discussion, the most challenging obstacle our group facing was ______ (please explain briefly).") targeting student's perceived challenges were analysed. These 291 free-text responses were coded into non-mutually exclusive dimensions of challenges: cognitive challenges (54%), metacognitive challenges (24%), emotional challenges (6%), and technical/other challenges (13%). A similar approach, 1 to 3 grams with TF-IDF features, was used to build a random forest classifier per dimension. Due to imbalanced data, especially in emotional challenges, an oversampling technique namely SMOTE (Synthetic minority over-sampling technique) was implemented. 5-fold cross-validation with stratified sampling was used to approximate the model performance. The models achieved average test accuracy and $F1_{weighted}$ of 0.80 and 0.78 respectively, improving to 0.82 and 0.80 with oversampling. This suggests acceptable classification models built from post-survey responses.

## 4.2 An expert knowledge model with a rule-based approach

As we want to explore the potential of engineered features in predicting the presence of challenges and further identifying a challenge dimension, a rule-based approach was formulated based on each binary feature. TF-IDF features were excluded in this approach because they aren't binary which require a classification threshold. Thus, TF-IDF features were only explored in the 2nd approach. For the remaining engineered features that are binary, we formed expert rules assuming that the presence of a selected feature may lead to a presence of challenge ('is_challenge'). These predicted presences of challenges were then forwarded to another set of rules experimenting with to what extent these features can be a good predictor of a specific challenge dimension ('is_cognitive_challenge', 'is_metacognitive_challenge', 'is_emotional_challenge', or 'is_technical/other_challenge'). It is worth noting that the engineered features were extracted at an utterance level and aggregated to be evaluated at an episode level. To illustrate, an episode that contains an utterance that was identified as a question ('is_question') will be evaluated to what extent this is aligned with the actual presence of challenges or a particular dimension of challenges in the episode. Evaluation results of each engineered feature in predicting challenges and their dimensions are presented using accuracy and $F1_{weighted}$ as quality measures.

## 4.3 A supervised machine learning model

The study utilized a human-coded dataset for supervised learning, employing two notable ML techniques: Support Vector Machine (SVM) and Random Forest (RF) classifiers. These were chosen due to their superior performance in classification tasks on a variety of datasets among hundreds of classifiers [16]. We experimented with TF-IDF features with/without additional engineered features compared across aggregation levels (utterances vs episodes). Through experimenting with parameters, RF classifiers with 100 estimators and 'balanced' class weights were deployed to encounter imbalance in the data class whereas a linear kernel was used for SVM. Both techniques were implemented using Python's Scikit-learn library. To follow the 2-step approach, we first performed 5-fold cross-validation with stratified sampling on a supervised classifier to predict the presence of a challenge. The test accuracy and $F1_{weighted}$ of each fold were then averaged and reported. Followingly, we built four independent classifiers, corresponding to four different dimensions of challenges, on the predicted data that was classified as challenges. The classifiers were evaluated on agreement of the predicted labels and the human labelled dimensions of challenges. Similarly, 5-fold cross-validation with stratified sampling was used for evaluation whereas average test accuracy and $F1_{weighted}$ were reported for model comparison. Feature importance from the RF classifier and coefficients from the SVM classifier were included for interpretation.

## 4.4 A Large Language Model Approach

In this study, GPT APIs were used through the OpenAI Python library to obtain paid services with advanced configuration from GPT. Unlike ChatGPT, GPT API provides opportunities to configure the GPT model and adjust some model parameters. GPT-4 was selected as an experimenting model in this study as it is the most advanced model that is available in the market at the time of the study. The model temperature or a sampling temperature ranging from 0 to 2 for the chat completion task where a lower value refers to "more focused and deterministic" outputs and a higher value refers to "more random" outputs[10]. In this case, we set the model temperature to zero as we used GPT for a targeted classification task, not a generative task that required higher diversity. We then constructed a prompt to retrieve a response from GPT.

The system role was set to a specific role namely "You are a teaching assistant observing students' discussion and helping teachers detect challenges in group discussion." For a prompt message, we initially provided it with a task context (e.g., "A group of students is working on a brainstorming task.."), followed by its task i.e. "Identify whether there are any challenges occurred and why." For the definition of types of challenges, we followed a concept from Xiao et. al., [56] by deploying codebook-centred prompts with few-shots. In their study, they investigated the performance of different prompt designs (Codebook-centred vs. Example-centred) using pre-trained LLM (GPT3) without fine-tuning in deductive coding tasks. While codebook-centred prompts mimic researchers' codebooks including descriptions of each code and its examples, example-centred prompts provide explanations of justification behind each example. The results show that codebook-centred prompts obtain fair to substantial agreement in coding with the experts' coding, higher than the example-centred prompts. Additionally, our codebook with examples was presented in detail to promote clarity in classification. In other words, the codebook was structured in a hierarchical format to reflect dimensions of challenges and their sub-dimensions (see Supplementary material S2[11]) as we assumed that the more context it has and the more distinctive the dimension is, the better the prediction will be. For example, while (C) refers to cognitive challenges, (C1) refers to a sub-dimension of cognitive challenge in expressing confusion either about the task, contents, or task expectation.

---

[10]https://platform.openai.com/docs/api-reference/chat/create
[11]https://osf.io/ta87g/?view_only$=$74d86897cf1a4b02808c5821221a3e42

Table 2: Performance results of the conventional NLP with a rule-based approach

| Feature | Is_challenge | | Is_cognitive_ challenge | | Is_emotional_ challenge | | Is_metacognitive_ challenge | | Is_technical/other_ challenge | |
| --- | --- | --- | --- | --- | --- | --- | --- | --- | --- | --- |
| | Accuracy | $F1_{weighted}$ | Accuracy | $F1_{weighted}$ | Accuracy | $F1_{weighted}$ | Accuracy | $F1_{weighted}$ | Accuracy | $F1_{weighted}$ |
| has_pron | 0.81 | 0.76 | 0.56 | 0.41 | 0.11 | 0.03 | 0.30 | 0.16 | 0.18 | 0.07 |
| has_pron & is_question | 0.75 | 0.77 | 0.62 | 0.48 | 0.12 | 0.04 | 0.33 | 0.19 | 0.19 | 0.08 |
| is_anger | 0.81 | 0.79 | 0.58 | 0.44 | 0.12 | 0.04 | 0.30 | 0.16 | 0.18 | 0.07 |
| is_annoyance | 0.80 | 0.79 | 0.58 | 0.44 | 0.11 | 0.04 | 0.30 | 0.16 | 0.19 | 0.08 |
| is_confusion | 0.83 | 0.80 | 0.59 | 0.45 | 0.11 | 0.03 | 0.31 | 0.16 | 0.18 | 0.07 |
| is_curiosity | 0.82 | 0.78 | 0.57 | 0.43 | 0.11 | 0.03 | 0.30 | 0.16 | 0.18 | 0.07 |
| is_disappointment | 0.74 | 0.76 | 0.61 | 0.47 | 0.13 | 0.04 | 0.32 | 0.17 | 0.20 | 0.08 |
| is_disapproval | 0.81 | 0.79 | 0.58 | 0.44 | 0.12 | 0.04 | 0.30 | 0.16 | 0.18 | 0.07 |
| is_disgust | 0.76 | 0.76 | 0.60 | 0.46 | 0.11 | 0.04 | 0.30 | 0.16 | 0.20 | 0.08 |
| is_embarrassment | 0.77 | 0.77 | 0.58 | 0.44 | 0.12 | 0.04 | 0.31 | 0.17 | 0.19 | 0.08 |
| is_fear | 0.73 | 0.76 | 0.62 | 0.48 | 0.12 | 0.04 | 0.30 | 0.17 | 0.21 | 0.09 |
| is_grief | 0.71 | 0.74 | 0.60 | 0.46 | 0.13 | 0.04 | 0.33 | 0.19 | 0.19 | 0.09 |
| is_nervousness | 0.74 | 0.75 | 0.60 | 0.46 | 0.12 | 0.04 | 0.32 | 0.18 | 0.18 | 0.07 |
| is_question | 0.80 | 0.80 | 0.61 | 0.48 | 0.12 | 0.04 | 0.31 | 0.16 | 0.19 | 0.08 |
| is_rejection | 0.58 | 0.63 | 0.65 | 0.53 | 0.14 | 0.05 | 0.31 | 0.18 | 0.21 | 0.10 |
| is_remorse | 0.48 | 0.53 | 0.61 | 0.48 | 0.17 | 0.08 | 0.35 | 0.21 | 0.27 | 0.14 |
| is_sadness | 0.55 | 0.60 | 0.60 | 0.47 | 0.16 | 0.06 | 0.35 | 0.22 | 0.25 | 0.13 |
| neg_sentiment | 0.65 | 0.70 | 0.66 | 0.54 | 0.16 | 0.07 | 0.32 | 0.17 | 0.22 | 0.10 |

We also asked it to formulate output responses in a certain format (e.g., "Provide your response in JSON format as follows."). Altogether, the constructed prompt helps hint them towards 2-step approaches in firstly identifying whether there is a challenge and secondly reasoning their responses pointing to a particular challenge that it can capture. Finally, each episode of discourse was attached to the prompt and sent to GPT API. We prepared episodic discourse by combining speaker information and utterances within an episode into a single text. As a result, GPT returned texts in JSON (JavaScript Object Notation) format for further processing. As we observed that the model tends to reason its decision using direct codes from the codebook (e.g., "(C1)"), a combination of regular expression and JSON parsing was utilised to post-process the returned data into structured formats, like DataFrame, to evaluate the model performance. Similarly, accuracy and $F1_{weighted}$ in predicting challenges, followed by a detected dimension of challenges are reported in the result section.

## 5 RESULTS AND DISCUSSION

Performance results from three different NLP modelling techniques (rule-based approach, supervised ML, and GPT4) in predicting students' challenge moments in collaborative discourses are presented in table 2, table 3, and table 4, accordingly. The results of predicting challenges, followed by identifying the dimensions of challenges will be discussed.

### 5.1 The expert knowledge model with a rule-based approach

In terms of detecting challenges, the rule-based approach received an average accuracy = 0.73, $F1_{weighted}$ = 0.74 across engineered features. The best-performed features in $F1_{weighted}$ are 'is_confusion' and 'is_question' which suggest that a challenge tends to happen when confusion or inquiry has been made. To our surprise, a combination of 'has_pron' and 'is_question' features did not perform better, compared to 'has_pron' or 'is_question' features. This indicates that a challenge happens when students refer to a pronoun or make an inquiry rather than a co-occurrence of the two events.

However, when further identified a dimension(s) of challenges, the rule-based model performed poorly with approximate scores across the engineered features lower than 0.5 for all dimensions, except identifying cognitive challenges that obtained an average accuracy of 0.60 ($F1_{weighted}$ = 0.46). Since this is a binary classification problem, a baseline lower than 0.5 would be considered lower than chance and unacceptable. Hence, while the rule-based model built from a single engineered feature might be useful in predicting a challenging moment, the model is impractical in determining the dimensions of challenges. This might be due to the low appearance of metacognitive (26%), technical/other (18%) and emotional (10%) challenges in the dataset compared to the high prevalence of cognitive challenges at 52%.

### 5.2 The supervised machine learning model approach

As we have multiple experiments within the second approach, the results of comparing the analysis at different units of analysis (utterance vs. episode level), the used features (TF-IDF vs. multiple features) and types of classifiers (RF and SVM), are included here. In general, the analysis at the utterance level performed slightly better than the episode level. To illustrate, when predicting whether there is a challenge, the average accuracy and $F1_{weighted}$ of the models across classifier types and features are 0.85 and 0.82, respectively whereas the model dropped its performance to an accuracy of 0.82 and $F1_{weighted}$ of 0.80 when analysing at episode level. This suggests that features extracted at the utterance level, especially TF-IDF features, can result in a better discriminating power in predicting a challenge than when aggregating at the episode level. To understand the different decisions made among the predictive models, the top-20 features identified from feature importance of RF or coefficients of SVM compared between pipelines were reported in Supplementary material S3[12]. We observed that the recognised features produced at the episode level tend to be more general in comparison to utterance-level features. For example, in predicting whether there is a challenge, the SVM classifier produced from TF-IDF features at episode level recognised keywords such as 'yeah', 'oh', and 'so' whereas the similar model analysed at utterance level acknowledged the importance of words such as 'what happen', 'what mean', and 'confus.' Please note that the reported words are preprocessed words (stemming and lemmatization) which might not be grammatically correct. The generic words that appear significant in classification models could show the limitation of using word frequency as a feature. Even with the TF-IDF weighted function, it seems that the models still attributed their weights using common yet frequent phrases rather than distinctive keywords. This also highlighted another limitation of the conventional NLP approach that does not perform well in indicating keywords when it is exposed to higher amounts of text i.e., aggregated data at episode level.

Considering the models with/without engineered features, the model performed better with additional features (accuracy = 0.84, $F1_{weighted}$ = 0.81) in addition to the existing TF-IDF features (accuracy = 0.83, $F1_{weighted}$ = 0.78) in the challenge prediction task. On the contrary, the challenge dimension detection models built from TF-IDF and engineered features performed equally at the average accuracy = 0.82, $F1_{weighted, TF-IDF}$ = 0.80 and $F1_{weighted, multifeatures}$ = 0.79. This suggests that engineered features may not notably enhance model performance which is consistent with previous research that feature engineering may not always result in effective models, particularly with complex target constructs [24].

For classification algorithms, RF and SVM showed comparable performance similar to [16]. RF slightly performed better in detecting a challenge (accuracy = 0.84, $F1_{weighted}$ = 0.81) than SVM (accuracy = 0.83, $F1_{weighted}$ = 0.78). On the contrary, SVM performed slightly better in determining a challenge dimension (accuracy = 0.83, $F1_{weighted}$ = 0.80) than RF (accuracy = 0.80, $F1_{weighted}$ = 0.79). Top features were then investigated. We found that the engineered features appeared as top features in RF but not in SVM. For example, 'is_question', 'neg_sentiment', 'is_curiosity' and 'is_confusion' were considered as RF's key features in predicting a challenge whereas TF-IDF features such as 'what happen', 'confus', 'do need' and 'what mean' were indicators captured by SVM. This might suggest that ensemble learning techniques (RF) tend to resemble human thinking by going beyond word occurrences to account for contextual meaning.

In general, the supervised classifiers, regardless of algorithms, built from multiple features at utterance level show high performance in indicating challenge(s) (accuracy = 0.85, $F1_{weighted}$ = 0.82), identifying cognitive challenge(s) (accuracy = 0.70, $F1_{weighted}$ = 0.70), emotional challenge(s) (accuracy = 0.95, $F1_{weighted}$ = 0.93), metacognitive challenge(s) (accuracy = 0.83, $F1_{weighted}$ = 0.79) and technical/other challenge(s) (accuracy = 0.81, $F1_{weighted}$ = 0.81). Referring to the top-rank features, words that are related to resources or concepts such as 'data', 'graph' and 'theori' appear as important features in predicting cognitive challenge(s) in the SVM classifier. There are also words that are related to confusion expression such as 'differ', 'what', 'mean', 'does make sens', 'whi one' appear in the SVM's top features. For the RF classifier, top features tend to be the engineered and TF-IDF features that target confusion (e.g., 'what', 'mean', 'differ', 'is_question', and 'is_confusion') and frustration expression (e.g., 'is_disgust', 'is_annoyance', and 'is_anger'). In predicting emotional challenge(s), emotional-related TF-IDF keywords such as 'sorri confus', 'brain', 'happen what', 'die', 'exhaust', appear prominently as important features in SVM whereas minor key features were the emotional-related engineered features (e.g., 'is_reject', 'is_sadness', 'is_remorse'). This is in the opposite of the top features generated from RF where emotional-related engineered features appear predominantly (e.g., 'is_fear', 'neg_sentiment', 'is_nervousness'). Key features derived from predicting metacognitive challenge(s) using SVM are highly related to the expression of confusion (e.g., 'still confus', 'how know relat'), confirmation of task execution (e.g., 'do group', 'anyth need', 'did', 'and whi') and controlling of tasks (e.g., 'finish', 'did finish', 'we finish right'). In contrary, top features generated from RF are mainly a combination of emotional and cognitive engineered features such as 'is_disgust', 'is_anger', 'is_question', 'neg_sentiment', 'is_curiosity' while a small proportion of TF-IDF keywords that refer to tasks (e.g., 'correl'), task understanding (e.g., 'what', 'whi') and task controlling (e.g., 'finish'), appeared significant. It is worth noting that time- or progress-related keywords didn't feature in the top 20 keywords as they rarely appeared in the dataset. At last, both RF and SVM shared their top features in predicting technical/other challenge(s) where a majority of keywords is related to resources or tasks such as 'whi go', 'whi background', 'neutral' (a contextual keyword from one of the tasks), 'duplic' (experienced technical problems in file duplicates), 'limit vote', 'ca make edit' (constraints of a collaborative platform), and 'sorri' (requesting a friend help) whereas a minority of keywords links to emotional engineered features such as 'is_anger', 'is_remorse', etc. However, none of the post-survey features appeared significant in any models. This is partly because there were limited numbers of responses resulting in low predictive power of the models compared to other engineered features.

### 5.3 The Large Language Model Approach

The classification performance performed by LLMs, in this case, GPT-4, is shown in Table 4. With minimal instruction and few-shot examples, GPT-4 performed exceptionally well in predicting a challenge(s) (accuracy = 0.83, $F1_{weighted}$ = 0.82) and identifying a challenge dimension (accuracy = 0.83, F1$_{weighted}$ = 0.83 across three dimensions), except identifying metacognitive challenges which have acceptable performance (accuracy = 0.65 and F1$_{weighted}$ = 0.66). In comparison to other dimensions of challenges, metacognitive challenges contain, most probably, highly different and contextual sub-dimensions which are M1: raising concerns over time/progress, M2: expressing confusion when executing tasks and M3: expressing concerns over group strategies/approaches. To illustrate, excerpt 1 shows metacognitive challenges of Daniel during task execution. Daniel acted as a facilitator of the group, asking a member of the team to present their ideas written on a sticky note. In this episode, he asked "Who's purple?" where 'purple' in this case refers to a sticky note's colours on the brainstorming platform. Without this contextual-specific information, the model might fail to understand how the purple colour is related to the task. Similarly, group strategies are also contextual-specific as illustrated in excerpt 2. In this case, Antony showed confusion in the group approach, so he started to question the 'limited vote' approach. Since multiple things could be recognised as 'approach' or 'strategy' in the contexts, the word 'approach' or 'strategy' itself might not actually appear in the contexts. This resonated with the findings emphasised in Zambrano et. al. [57]'s study that GPT-4 (through ChatGPT in their case) tends to follow its pre-existing knowledge and may produce overgeneralised results when differentiating constructs that are subject to interpretation or ambiguous. Therefore, we hypothesised that the model might benefit from extra contexts provided e.g., What is the task? What happens during task execution? What are the possible approaches? Also, additional clarification on the definition of subdimensions and extra examples may also be helpful in training the model. Apart from this, an inclusion of justification on why a particular example was classified as a certain category, can also be experimented [56]. However, this was not considered in the scope of the present study.

---
[12]https://osf.io/ta87g/?view_only$=$74d86897cf1a4b02808c5821221a3e42

Table 3: Performance results of the conventional NLP with supervised machine learning

| Approach | Is_challenge | | Is_cognitive _challenge | | Is_emotional _challenge | | Is_metacognitive _challenge | | Is_technical/other _challenge | |
|---|---|---|---|---|---|---|---|---|---|---|
| | Accuracy | F1$_{weighted}$ | Accuracy | F1$_{weighted}$ | Accuracy | F1$_{weighted}$ | Accuracy | F1$_{weighted}$ | Accuracy | F1$_{weighted}$ |
| Utterance | 0.86 | 0.82 | 0.73 | 0.72 | 0.98 | 0.97 | 0.86 | 0.86 | 0.83 | 0.86 |
| TF-IDF features | 0.86 | 0.82 | 0.75 | 0.73 | 0.97 | 0.96 | 0.83 | 0.84 | 0.85 | 0.87 |
| RF | 0.85 | 0.82 | 0.78 | 0.77 | 0.99 | 0.99 | 0.77 | 0.82 | 0.76 | 0.83 |
| SVM | 0.87 | 0.82 | 0.72 | 0.70 | 0.95 | 0.93 | 0.89 | 0.86 | 0.94 | 0.91 |
| Multifeatures | 0.87 | 0.82 | 0.72 | 0.70 | 0.98 | 0.98 | 0.89 | 0.88 | 0.81 | 0.84 |
| RF | 0.86 | 0.82 | 0.76 | 0.75 | 0.99 | 0.99 | 0.91 | 0.91 | 0.68 | 0.78 |
| SVM | 0.87 | 0.83 | 0.67 | 0.66 | 0.97 | 0.96 | 0.87 | 0.85 | 0.94 | 0.91 |
| Episode | 0.81 | 0.77 | 0.69 | 0.69 | 0.89 | 0.84 | 0.73 | 0.65 | 0.82 | 0.76 |
| TF-IDF features | 0.80 | 0.75 | 0.70 | 0.70 | 0.89 | 0.84 | 0.73 | 0.65 | 0.83 | 0.77 |
| RF | 0.81 | 0.79 | 0.70 | 0.70 | 0.89 | 0.83 | 0.72 | 0.61 | 0.82 | 0.74 |
| SVM | 0.79 | 0.70 | 0.71 | 0.70 | 0.89 | 0.85 | 0.75 | 0.70 | 0.84 | 0.80 |
| Multifeatures | 0.81 | 0.79 | 0.68 | 0.68 | 0.89 | 0.84 | 0.73 | 0.65 | 0.82 | 0.75 |
| RF | 0.82 | 0.80 | 0.68 | 0.68 | 0.89 | 0.83 | 0.72 | 0.61 | 0.82 | 0.74 |
| SVM | 0.81 | 0.78 | 0.69 | 0.68 | 0.89 | 0.84 | 0.75 | 0.68 | 0.82 | 0.76 |

| | | | | |
|---|---|---|---|---|
| Excerpt 1 | | | Excerpt 2 | |
| Daniel: | Who's purple? | | Anthony: | So do you have limited votes? |
| Daniel: | So the purple. | | Daniel: | Oh, yeah. |
| Anthony: | I'm the purple. | | Anthony: | For how many? |
| Daniel: | You're the purple, okay. | | Daniel: | Yeah, you can have a first and a second choice. |

In addition to presenting the model's performance, we are unable to probe into the underlying factors influencing the model's decision-making process, as LLMs use a black-box approach. Even though we asked the model to output its justification, this may not be equivalent to model weights returning from the supervised ML approach. It is worth being cautious when referring to the formulated rationale as this is a probabilistic-based generated content from the model.

In comparisons, the two approaches namely NLP with supervised ML and LLMs performed exceptionally well by receiving average accuracy and F1$_{weighted}$ above 0.70 across tasks (except in identifying metacognitive challenges from the GPT-4 model). This suggests the potential of the supervised ML approach and the advanced LLMs in predicting students' challenge moments and their dimensions in collaborative discourses. On the other hand, a simple rule-based approach mapping the engineered features with the targeted constructs may be overly simplistic to accurately capture such complex learning constructs as dimensions of challenges.

### 5.4 The relative advantages and disadvantages of models for improving teachers' practice

Since our goal is not necessarily to improve the state of the art in NLP but to improve educational practice with applied AI, we also considered the relative advantages and disadvantages of the three approaches for teaching and learning, particularly for their potential

Table 4: Performance results of GPT-4

| Step | Predicted Labels | Accuracy | F1$_{weighted}$ |
| --- | --- | --- | --- |
| 1 | Is_challenge | 0.83 | 0.82 |
| 2 | Is_cognitive_challenge | 0.76 | 0.75 |
|  | Is_emotional_challenge | 0.89 | 0.89 |
|  | Is_metacognitive_challenge | 0.65 | 0.66 |
|  | Is_technical/other_challenge | 0.85 | 0.85 |

to provide meaningful feedback to students and/or teachers to close the feedback loop.

For rule-based and supervised ML models, feature importance and keywords could give direct evidence of the groups' challenges and regulation processes in the form of lists of keywords, word clouds, a network graph of keywords etc. Praharaj et al. [42] proposed a collaboration analytics approach visualising the richness of group conversation through a network of related keywords across pre-assigned roles. A similar approach, i.e., a network graph of keywords representing detected challenges can be applied to contextualise challenges in students' collaboration to support awareness among teachers and learners. Not only could this promote trustworthiness by helping learners and educators understand the rationales behind the models, but it could also provide an additional layer of visualisations of the ideas being discussed in the groups and help identify common challenges. This would be particularly useful for helping teachers decide whether to give whole-class feedback rather than providing individualised support to each group to address a commonly faced challenge in practice. With Järvelä et al., [27] suggesting that low SSRL groups fail to use deep-level strategies to regulate their CL process, it could also be valuable to further develop the models to be able to identify the proportion of deep-level vs. routine-level SSRL strategies used by each group to help recognise the groups that may need additional support in SSRL.

On the other hand, the LLM only provides a brief summary of the challenges faced by group members with few unreliable instances of evidence given for its decision-making process. For instance, "Karen struggles to understand the term 'chatbot' (C4). She also has difficulty expressing her idea clearly, *as seen when she says 'A what?' and 'What?'* (C3). Lisa and Brian also struggle to understand Karen's idea (C4)." (Session 1 - episode 6) [emphasis added]. Although, this seems to be a reasonable challenge identified by the LLM, the audio transcription tells another story. While there was a short-lived confusion around the term, 'chatbot', the main idea being expressed by Karen was consequentially overlooked, which contained much more valuable insights to their discussion: "It is challenging that [teachers] cannot take care of all the students. So, technology can help them to take care of them" (Session 1 - line 64). The indication of a challenge is not merely enough to evoke effective regulation - students and teachers alike need to acknowledge the real source of these challenges to activate the appropriate SSRL activities to overcome them [29]. Unfortunately, given the 'black box' nature of LLM decision-making processes, users are unlikely to be able to validate or rationalise the model's outputs, lowering the trustworthiness of the system, particularly when the LLM fails to perceive the challenges accurately.

Besides model performance and feedback opportunities, there are other considerations such as the development time of the model, usability/accessibility, cost, reliability, stability, generalisability/scalability and others, to be taken into account when AI techniques are applied in education. A rule-based and supervised ML approach requires higher development time and domain-specific expert knowledge to engineer features for the models to achieve satisfactory models. In contrast, LLMs, which have been pre-trained, require only minimal effort/prompts to achieve similar results, thus, their development time is notably shorter. This also links to another key advantage of LLMs which is their high accessibility. LLMs require no prior background in programming to create a model since the prompt is written in human language whereas the rule-based and supervised ML approaches require designers to construct a model using programming language. In addition, the current LLM that we used, GPT-4, requires a third-party paid service, unlike the rule-based/supervised ML approaches that can be set up freely. There might be open-access models with equivalent or better performance available in the near future, yet do not exist so far. Moreover, when requesting model-as-a-service, commercial LLMs may be considered unstable as corporate developers have full control over any changes in the models [8] in comparison to the other two approaches which also raises significant issues about privacy and ethics of data ownership. Moreover, LLMs are largely black-boxes where we only have access to their inputs and outputs. This makes the models less trustworthy in comparison to other approaches where we can inspect their rules/decisions. Also, similar outputs can be expected from the same inputs in the rule-based/supervised ML approaches whereas this is not the case for generative LLMs as they are stochastic models. Therefore, even with the same prompt, multiple executions could result in different outputs from LLMs. Finally, one main drawback of the traditional NLP approaches is the domain specificity of the lexical corpus. In other words, NLP tends to be tied to the terminologies presented in the contexts resulting in low generalisability of the model. However, LLMs gain advantages over this issue through the utilisation of a 'Large' corpus to pre-train the model. This can mitigate the model generalisation problem to a certain extent. Hence, LLMs show high applicability to perform a task even in a specific domain and have a higher capability to generalise into other contexts than the rule-based/supervised ML approaches.

Table 5 summarises the advantages and disadvantages of the three approaches per consideration discussed above. It is worth noting that which method should be used predominantly depends on the use case, goals and expertise of the designers/users as well as multiple social and ethical considerations. For example, a novice teacher with no programming background who wants to set up a short-term analytics system to assist students during collaboration, could consider constructing prompts for LLMs to perform the task. On the other hand, an expert in programming who wants to deploy long-term learning analytics to study students' struggling moments and provide feedback on their patterns, might consider deploying the rule-base or the ML approaches. It is also essential to point out that this is not a mutually exclusive approach where users have to select one, not the others but rather they can experiment with different approaches and justify what is best for which task and why to further complement the model advantages in particular settings.

Table 5: Comparison of the advantages and disadvantages of models across several considerations

| Consideration/Approach | Rule-base model | Supervised ML with white-box approach | Large language models |
|---|---|---|---|
| Performance | (-) Low performance | | (+) Medium to high performance |
| Feedback opportunities, transparency of the model | (+) Since it is a white-box model, features/rules can be utilised to give specific feedback to learners and teachers. | | (-) It is a black box model which makes it difficult to give feedback on its rationales (no reliability of the chain of thought explanations from the model) |
| Development time | (-) It is quite time-consuming to engineer features. It also required expert knowledge in making connections between data and targeted constructs. | | (+) Very low development time because a prompt is already written in human language. |
| Usability/Accessibility | (-) Required programming skills | | (+) Easy to access, no programming background is required |
| Cost | (+) Free | | (-) Commercial products [a] |
| Reliability | (+) High reliability since the model will give the same outputs for the same inputs. | | (-) Low reliability. Even with the same prompt, several executions could produce different outputs. |
| Stability | (+) Quite stable because users have full control over the model. | | (-) Given that the model is commercial, it might be updated. |
| Generalisability/Scalability | (-) Domain-specific, likely to over-fit, low generalisability | | (+) Higher potentials of generalisation |
| Ethics and privacy | (-) Concerns over bias and fairness of training dataset | | (-) High concerns over obscured data storage and copyrights. |

[a] the best-performing models we used (GPT-4) are currently commercial, which might change in the future with open-access models' increased performance.

## 6 LIMITATIONS AND FUTURE WORK

There exist some limitations within this research. First, as this study was conducted with a small sample size in a specific context and tasks, the result may not be generalised into other contexts. Even though we are expecting the LLMs approach to perform equally well in predicting challenge moments in other collaborative contexts, we haven't validated this since it is out of scope. For the rule-based/supervised ML approaches, we acknowledged the weakness of NLP features that are highly contextual which may not be applicable to other settings. Hence, additional research is needed to explore meaningful NLP features in their own contexts. Second, this study focused on a single modality, audio, to infer challenge moments from group collaboration whereas collaboration is embodied in physical contexts where students interact multidimensionally. Hence, challenge moments may be surfaced through other channels such as facial expressions, gestures or behavioural logs which were not included in this study. Integration of multimodal data might benefit in holistically capturing students' challenge moments [9]. Third, there was a technical challenge in collecting clean data for analysis. Since the study was conducted in ecologically valid settings i.e., multiple groups of students worked simultaneously in the shared physical space, it is challenging to retrieve high-quality audio data for a downstream task which is the main reason for the dataset being excluded. Another technical limitation is in automatically and accurately generating speaker-embedded transcriptions from audios which currently require human supervision which might hinder the opportunity to give timely feedback to students and teachers. Resolving this issue is probably out of our scope but we believe that this issue will be mitigated through technological progress. Lastly, we acknowledged that we only presented one possible conceptual idea of how feedback could be designed from the models but we have not investigated the value of feedback from these NLP models to teachers and learners. Our future work will be a validation study of our design proposal with actual stakeholders to investigate the extent to which meaningful and actionable feedback generated from the models can help improve learning and teaching practices.

## 7 CONCLUSION

In this study, student discourse obtained from a semester-long ecologically valid physical classroom was analysed. We constructed and compared the predictive performance of the three NLP models in detecting challenge moments and their dimensions during group collaboration. The results show that the supervised ML and the LLM approaches show equivalently high performance across most tasks except identifying *metacognitive challenges* which was too difficult for the LLM. In contrast, the rule-based approach performed poorly in classifying challenge dimensions. Beyond performance comparisons, the models' advantages and disadvantages in different aspects were also discussed, especially in their potential to generate feedback to help inform learning and teaching practices. Significant limitations of all approaches, including currently hyped LLMs, are highlighted. This work contributes to the broader field of learning analytics through a comparison of NLP models and a reflection on their potential for analytics feedback in improving collaboration.

## ACKNOWLEDGMENTS

We would like to thank 2022/2023 DUTE students in MA EdTech at UCL for granting permission to collect data for this study. This work was supported by UCL AI Co-creator projects (summer 2023).